\documentclass[journal]{IEEEtran}
\ifCLASSINFOpdf
\else
\fi

\usepackage{times}
\usepackage{epsfig}
\usepackage{graphicx}
\usepackage{amsmath}
\usepackage{amssymb}

\usepackage{float}
\usepackage{graphicx}
\usepackage{amsmath}
\usepackage{amssymb}
\usepackage{amsfonts}
\usepackage{graphicx}
\usepackage{amsmath}
\usepackage{amssymb}
\usepackage{marvosym}
\usepackage{bm}
\usepackage{subfigure}
\usepackage{multirow}
\usepackage{makecell}
\usepackage{latexsym}
\usepackage{hyperref}
\usepackage{color}
\usepackage{svg}
\usepackage{algorithm}
\usepackage{algpseudocode}
\begin{document}
%
\title{Self-Adaptive Partial Domain Adaptation}
%
%
%

\author{Jian Hu, Hongya Tuo, Shizhao Zhang, Chao Wang, Haowen Zhong, Zhikang Zou\\Zhongliang Jing, Henry Leung,~\IEEEmembership{Fellow, IEEE}, Ruping Zou

\thanks{Jian Hu, Hongya Tuo,Shizhao Zhang, Zhongliang Jing are with the School of Aeronautics and Astronautics, Shanghai Jiao Tong University, China, Chao Wang is with Pingduoduo Group, China. Haowen Zhong is with Zhejiang Lab, China. Henry Leung is with Department of Electrical and Software Engineering, University of Calgary, Canada. Ruping Zou is with the Third Academy of China Ordinance, China}
\thanks{Hongya Tuo is the corresponding author}
}

%
%

\markboth{Journal of \LaTeX\ Class Files,~Vol.~14, No.~8, August~2015}%
{Shell \MakeLowercase{\textit{et al.}}: Bare Demo of IEEEtran.cls for IEEE Journals}
%



\maketitle

\begin{abstract}
Partial Domain adaptation (PDA) aims to solve a more practical cross-domain learning problem that assumes target label space is a subset of source label space. However, the mismatched label space causes significant negative transfer. A traditional solution is using soft weights to increase weights of source shared domain and reduce those of source outlier domain. But it still learns features of outliers and leads to negative immigration. The other mainstream idea is to distinguish source domain into shared and outlier parts by hard binary weights, while it is unavailable to correct the tangled shared and outlier classes. In this paper, we propose an end-to-end Self-Adaptive Partial Domain Adaptation(SAPDA) Network. Class weights evaluation mechanism is introduced to dynamically self-rectify the weights of shared, outlier and confused classes, thus the higher confidence samples have the more sufficient weights. Meanwhile it can eliminate the negative transfer caused by the mismatching of label space greatly. Moreover, our strategy can efficiently measure the transferability of samples in a broader sense, so that our method can achieve competitive results on unsupervised DA task likewise. A large number of experiments on multiple benchmarks have demonstrated the effectiveness of our SAPDA.
\end{abstract}


%
\IEEEpeerreviewmaketitle

\maketitle

\IEEEdisplaynontitleabstractindextext

%
\IEEEpeerreviewmaketitle

{\section{Introduction}\label{sec:introduction}}

\IEEEPARstart{T}{ransfer} learning concentrates on knowledge transfer learned on labelled source domain to unlabeled target domain by narrowing the distance among source and target domains.~\cite{Hoffman18}~\cite{Saenko10}~\cite{wang2019domain} As one of the core approaches for transfer learning, domain adaptation aims to relieve the need for labelled data. ~\cite{Yosinski14}~\cite{li2018heterogeneous}~\cite{zhang2021domain}. There are two classes of methods to deal with it. The first is maximum mean discrepancy (MMD) which manages to maximize the distance between the mean of the two domains in some high dimensional mapping space. The second is the adversarial learning method, which attempts to extract the domain invariant features among the domains. Here, a basic assumption for domain adaptation is source and target domain share the same label space.~\cite{Pan19}~\cite{hu2020unsupervised} Whereas, this assumption is hard to satisfy in practice. 

Recently, partial domain adaptation task has been projected. Different from standard domain adaptation, PDA assumes that source label space contains target label space. This assumption relaxes the constraint of standard DA method that two domains have the same label space. Different from standard domain adaptation(Standard DA), PDA enables the transfer of knowledge from a domain with plenty of labels to another without labels, where source label space contains target one. Since large-scale annotated datasets such as Google
Open Images~\cite{goo} and ImageNet-1k~\cite{Russakovsky} are available, PDA can be applied to many practical applications. 

Partial domain adaptation task contains two crucial points. Firstly, compared with standard DA, PDA is more challenging because target label space is a subset of source label space~\cite{Jian2} and is unknown. The common subset of label space between different domains is denoted as shared domain while the part only including source label space is defined as outlier domain. During training of the PDA, an important issue is to select samples belonging to the shared domain. This dilemma can cause negative transfer if we misclassify some classes in the outlier domain to the shared one. Under this circumstance, some target samples can be wrongly grouped into these outlier domain. Secondly, the gap between source shared and target domain should be narrowed during training. Thus, we need to promote a universal framework to deal with it. Some previous methods  ~\cite{Cao18}~\cite{Cao181}~\cite{Cao19}~\cite{Zhang18}~\cite{Jian1} are proposed to handle PDA by weighing classes or samples in a domain adversarial network. They increase weights of the shared classes while decrease weights of outlier classes. In ideal situations, weights of the shared classes should be 1, and those of the outlier classes should be 0. However, they all use  probability-weights as class weights. Even though the weights of the source shared classes are meaningfully higher than the source outlier's, they are far away from 1. As a result, negative transfer still can not be avoided in these methods. Moreover, when the discrepancy among different domains are huge, the boundary between the shared and outlier domain is not clear. So if we just cluster source domain into two groups named as shared and outlier domains, some samples on the boundary are easily to be misclassified, which may lead to aligning features from target domain to outlier domain in the following training steps. As a consequence, it will result in performance degradation. 
\begin{figure*}[t]
\centering
\includegraphics[scale=0.4]{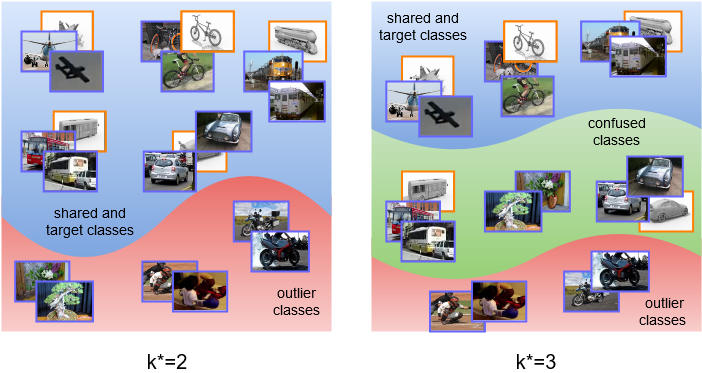}
\caption{Our self-adaptive partial domain adaptation mechanism, the source samples are in purple border and the target samples are in orange. If the best cluster number $k^*=3$, $w_i$ are obtained by the means of the class probability-weights in the group. Source and target samples are clustered into three groups by confidence: Groups with highest confidence(weights are close to 1 or 0) are shared and outlier classes respectively, while the group with a low degree of confidence(weights are neither close to 0 nor 1) is group with intermediate classes. In this way, The easily distinguishable shared and outlier classes are separated respectively, while the confused shared and outlier classes are grouped together. As thus, the distance between the high-confidence source shared and the target samples can be effectively narrowed, and the negative transfer caused by the misjudgment of the confused class can be prevented. If the best cluster number $k^*=2$, source shared and target samples are clustered into shared groups, and source outlier samples are clustered into outlier group. $w_i$ in shared group are set as 1, in outlier group are set as 0.}
\label{fig:universe1}
\end{figure*}
To copy with the difficulties of partial domain adaptation, an end to end Self-Adaptive Partial Domain Adaptation (SAPDA) network is proposed. Previous PDA methods only cluster source classes into shared and outlier classes, but the samples on the boundary can be easily misclassified when the discrepancy between different domains is huge, which can cause negative transfer greatly. To deal with this problem, in this paper, SAPDA self-adaptively clustered source label space into different groups. When the number of group is three, our model can not only select out the high-confidence samples as shared and outlier classes, but can also put the samples that are difficult to distinguish into the confused classes. In this way, when the shared and outlier classes are not easy to distinguish, our model can effectively narrow the general gap between high-confident source shared classes and target classes while ignoring the effects of confused classes temporarily. If the groups are only two, SAPDA considered that it has enough confidence to select all the shared and outlier classes clearly. Samples in the same group share the same weights. The higher the weight is, the more possible the sample belongs to shared domain. Meanwhile, the self-adaptive weighted mechanism updates each 500 iterations, which can dynamically correct incorrect clustering weights. In this way, we can avoid misclassification due to the entanglement between shared and outlier domain, and eliminate negative transfer greatly.

 SAPDA builds weighted adversarial network to straiten the gap between weighted source and target features, and reduce negative transfer of outlier domain. We carry out comprehensive experiments on five representative domain adaptation benchmark datasets.

\begin{figure*}[t]
\centering
\includegraphics[scale=0.46]{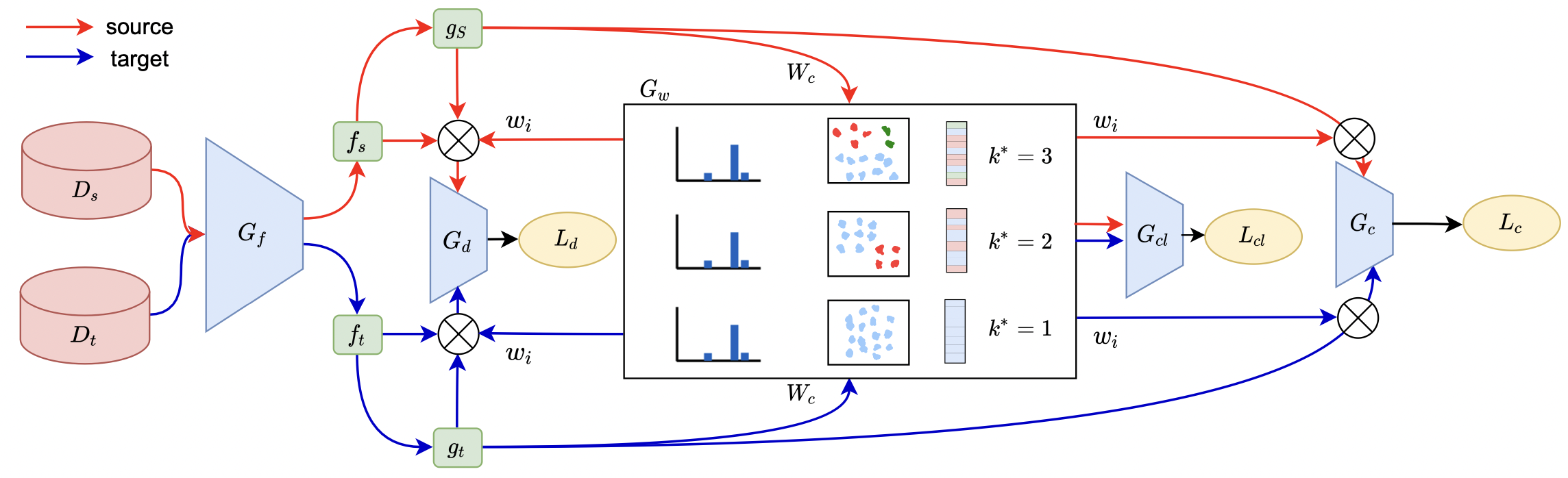}
\caption{Our SAPDA framework. $G_f$ is the feature extractor, $G_d$ is the domain discriminator, $G_c$ is the source classifier, $W_c$ is the class weight vector. $G_w$ is the self-adaptive class weights evaluation mechanism, $w_i$ is the class weight.  $G_{cl}$ is the cluster classifier. The red and blue flows are from source and target domains respectively.}
\label{fig:universe}
\end{figure*}

\section{Related Work}\label{sec:related work}
\subsection{Domain Adaptation}
Domain adaptation is an approach that tries to build domain invariance between different domains, and mitigates the burden of annotating target data. A recent study has indicated that deep neural networks can learn invariant representations. These invariant representations can help knowledge transfer between domains.

Even if deep neural networks can disentangle complex data distributions, the discrepancy across domains can not be removed. Hence, recent works focus on how to connect deep neural network and domain adaptation. Two main approaches are proposed to handle it. The first approach tries to match the high order statistic features by adding an adaptation layer~\cite{inproceedings}~\cite{Long15} ~\cite{Long16}~\cite{Wang2019}~\cite{Li}. The second one tries to extract common features cross domains by appending a domain discriminator ~\cite{Ganin16}~\cite{Tzeng15} ~\cite{zhong2019transfer}~\cite{Tzeng17}~\cite{Luo}.

Recently, how to combine domain adaptation to realistic application has also get more and more attention. Domain adaptation has been designed as an universal module, applying to object detection~\cite{2018Pedestrian} ~\cite{2018Domain} ~\cite{Song20}, semantic segmentation ~\cite{2020Pan} ~\cite{2019ADVENT} and person re-id~\cite{zhao2020unsupervised} ~\cite{Yang}. They have made a great contribution to alleviate the lack of labels in practical application.

\subsection{Partial Domain Adaptation} Partial domain adaptation assumes target label space is a subset of source label space~\cite{Jian2}. Some methods have been presented to deal with PDA problems. Selective
Adversarial Network (SAN)~\cite{Cao181} employs numerous adversarial networks and relative importance mechanisms to filter outlier classes. Partial Adversarial Domain Adaptation (PADA)~\cite{Cao18} modifies SAN by adding class-level relative importance index to source classifier to build a general adversarial network. Importance Weighted Adversarial Nets (IWAN)~\cite{Zhang18} utilizes an attached domain discriminator to evaluate the sample-level weight and further adds the weight to adversarial network. Example Transfer Network (ETN)~\cite{Cao19} uses an auxiliary adversarial network to evaluate sample-level weight and to add the weight on adversarial network. These approaches can effectively process PDA compared with traditional methods. 

These methods all use probability-weights to find out whether the class belongs to shared classes. However, in practice, if one class is subject to shared classes, the weight hardly achieves 1. This phenomenon can cause negative transfer. Moreover, when the discrepancy between domains is huge, the boundary between shared and outlier is not clear. So it is hard to classify the samples on the boundary. In other words, if we divide source domain into shared domain and outlier one directly, some of them are easy to misclassify. This paper proposes Self Adaptive Partial Domain Adaptation (SAPDA) that clusters source classes automatically into several different groups according to class probability-weights. Samples in the same group share the same weights. If there are more than two groups, sample weights are obtained by the means of the class probability-weights in the group. Specially, if the groups are only two, sample weights are given as 1 or 0.  Furthermore, we utilize Calinsk-Harabaze index ~\cite{in} to evaluate the optimal number of groups we cluster into. In this way, we can guarantee a good classification.
\section{Self-Adaptive Partial Domain Adaptation}

\subsection{Preliminaries}
In PDA, we define $C_s$ as the source label space, $C_t$ as the target label space. Source domain is defined as $\mathcal{D}_s = \{(\mathbf{x}_i^s,y^s_i)\}_{i=1}^{n_s}$ of $n_s$ labeled samples with $|C_s|$ classes, while target domain is defined as ${{\mathcal D}_t} = \{ {\mathbf{x}}_i^t\} _{i = 1}^{{n_t}}$ of $n_t$ unlabeled samples with $|C_t|$ classes. Under the PDA setting, $C_s$ contains $C_t$. Source label $C_s$ consists of two parts, shared label space $C_{sh}$ and outlier one $C_{so}$. $C_{sh}$ is the same as target label space $C_t$, which means they have the same set of label classes. Outlier label space $C_{so}$ is the unique part of $C_{s}$, that to say, $C_{so}$ = $C_{s} \setminus C_{sh}$. ${\mathcal D}_{sh}$ is source-shared domain and ${\mathcal D}_{so}$ is source-outlier domain.

Different from previous PDA methods, we introduce ${\mathcal D}_{conf}$ as the domain of samples that are hard to distinguish. On the one hand, if our novel self-adaptive weight mechanism can classify all the source sample into source shared and outlier classes confidently, then we have ${\mathcal D}_{s}= {\mathcal D}_{so}\cup{\mathcal D}_{sh}$. On the other hand, if our model is not confident enough to clearly distinguish the samples around the border, we cluster these confused samples into a separate cluster. Under this circumstance, ${\mathcal D}_{s} = {\mathcal D}_{so}\cup{\mathcal D}_{sh}\cup{\mathcal D}_{conf}$. Our self-adaptive weight updates each 500 iterations.

We assume that source domain ${\mathcal D}_s$ and target domain ${\mathcal D}_t$ are sampled from distributions $p_s$ and $p_t$ respectively. Similarly, $p_{sh}$ denotes the distribution of source-shared domain ${\mathcal D}_{sh}$ and $p_{t}$ denotes the distribution of source-outlier domain ${\mathcal D}_t$.

In PDA task, the key issue includes two parts: firstly, due to shared and outlier classes are unknown in advance, it is essential to select out unrelated source data belonging to ${\mathcal D}_{so}$ in order to decrease negative transfer. Secondly, we will try to narrow the gap between ${\mathcal D}_{sh}$ and ${\mathcal D}_t$. These two issues should be dealt simultaneously.

Our architecture of SAPDA is shown in Fig.\ref{fig:universe}. Class weight $w_i$ is the core variable of our proposed method. It denotes the probability that the sample $x_i$ comes from the source shared domain. Samples in the same group of classes (either in shared group, outlier group or confused group) share the same $w_i$. $w_i$ acts on some modules, including the domain discriminator $G_d$, the source classifier $G_c$ and the cluster classifier $G_{cl}$. It aims to help the network focus on shared domain samples and exclude outlier samples. In the following subsections, we will introduce each modules of the model and how $w_i$ help them to focus on shared domain samples.
 
In subsection B, we introduce the basic building blocks of our model, including feature extractor $G_f$, $G_c$, $G_d$ and $G_{cl}$.

In subsection C, we specify how the self-adaptive class weights evaluation mechanism $G_w$ computes $w_i$. In a nutshell, $G_w$ takes $W_c$ as input and outputs $w_i$. $W_c$ is a $|C_s|$-dimension vector. It is the average of outputs from source classifier $G_c$ of all target samples. That is, $W_c=\frac{1}{n_t}\sum_{i=1}^{n_t}(G_c(G_f(x_i^t)))$. $W_{c}[j]$ represents the possibility of the $j$-th source domain class being part of the shared classes $C_{sh}$. We use $W_c$ to compute $w_i$.

In subsection D, we explore how $w_i$ acts on $G_c$ and $G_d$ as a weight value. The influence of $w_i$ is that $G_c$ and $G_d$ is guided to focus on the shared domain samples and not the outlier ones. In this way, positive transfer contributed by samples in $D_{sh}$ is enhanced and negative transfer caused by samples in $D_{so}$ is mitigated.

In subsection E, we explain why and how $w_i$ is used as cluster label for cluster classifier $G_{cl}$.

\subsection{Basic building blocks}
Feature extractor $G_f$ extracts features $f_i$ from input image samples $x_i$. That is, $f_i = G_f(x_i)$.

Source classifier $G_c$ takes extracted features as input and outputs $|C_s|$-dimension vectors $g_{i}$. That is, $g_{i}=G_c(G_f(x_i))$ and $|g_i|=|C_s|$. $g_{i}[j]$ represents the possibility of sample $x_i$ belonging to the $j$-th source domain classes. The loss of $G_c$ is denoted as:

\begin{equation}
\begin{split}
    L_{c}(G_c)=&\frac { 1 } { n _ { s } } \sum _ { \mathbf i=1 }^{n_s} L  \left( G_c\left( G_f  \left( \mathbf { x } _ { i }^s \right) \right) , y _ { i } \right)
\end{split}
\end{equation}

Domain discriminator $G_d$ plays the minimax game with $G_f$ to extract domain-invariant features. $G_d$ tries to tell the difference between source domain samples and target domain samples using the features given by $G_f$ while $G_f$ tries to generate domain-agnostic features to confuse $G_d$. This adversarial interaction between $G_f$ and $G_d$ leads to domain-invariant features. As a result, the source classifier $G_c$ can be applied to the task in $D_t$ instead of just $D_s$. This framework is proposed by ~\cite{Ganin16} and it utilizes a minimax loss to implement adversarial weighted domain adaptation:

\begin{equation}
\begin{split}
\min_{G_{f}} \max_{G_{d}} {L_d}(G_{f}, G_{d}) & = \mathbb { E } _ { x  \sim p _ { s } } \left[log \left( G_d ( G_f\left ( x \right ) \right) ) \right]\\& + \mathbb { E } _ { x \sim p _ { t } } [  log\left(1- G_d \left( G_f\left ( x \right ) \right) ) \right] 
\end{split}
\end{equation}

Cluster classifier $G_{cl}$ self-adaptively divides source domain into several groups, including shared classes group, outlier classes group and confused classes group. By minimizing the loss of $G_{cl}$, samples within the same group will be pulled together and samples from different groups will be pushed away from each other. As a result, the model can better classify target domain samples (which belong to the shared classes) without the interference of by outlier samples.

\begin{figure}[t]
\centering
\includegraphics[width=\linewidth]{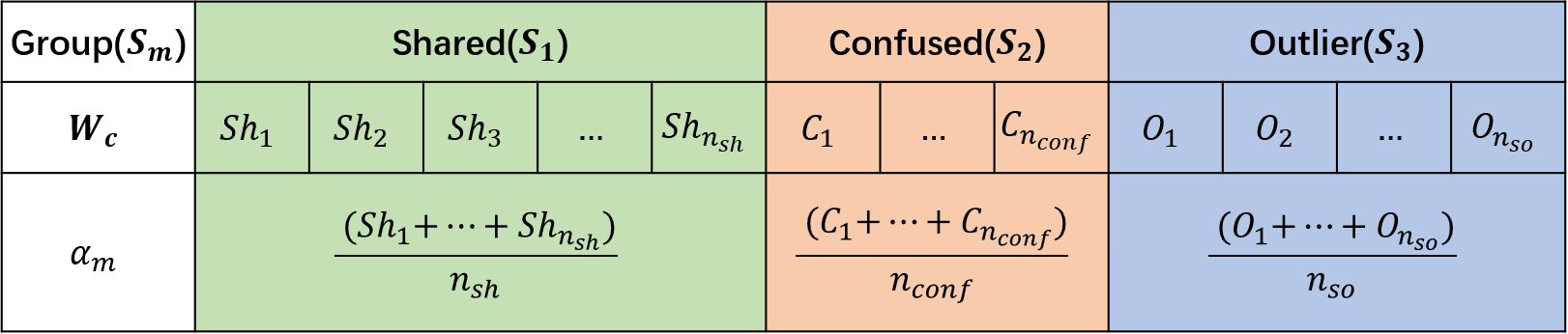}
\caption{Illustration of different variables}
\label{fig:varrela}
\end{figure}

\subsection{Self-Adaptive Class Weights Evaluation}

To compute class weights $w_i$, we need to first decide which classes of the source label space are the shared classes.  Similar to PADA~\cite{Cao18}, we use the aforementioned weight vector $W_c$ to make this decision. By averaging the outputs from source classifier of all target samples ${\{\mathbf{x}}_i^t\} _{i = 1}^{{n_t}}$, $W_c$ characterize the probability  of each source class being the shared classes. Because a basic assumption in PDA is that $p_t$ is much more similar to $p_{sh}$ compared with $p_{so}$.

We normalize $W_c$ to avoid some elements being too small:
\begin{align}
    W_c=\frac{W_c}{max(W_c)}
\end{align}

Ideally, weight values in $W_c$ for shared classes ${C}_{sh}$ should be 1 and those for outlier classes ${C}_{so}$ should be 0. But in reality neither of them can achieve the ideal values, which leads to negative transfer. To solve this problem, a basic idea is to set a reasonable threshold. If the $j$-th element $W_c[j]$ is higher than the threshold, the corresponding class weight should be converted to 1, otherwise 0. 

But this scheme has two difficulties: first, $W_c$ updates every few iterations, so it is hard to give a constant threshold. Second, when the domain discrepancy is huge, the boundary between shared and outlier domain is not clear. If we rashly classify samples into shared or outlier domains, misclassification can lead to performance degeneration.

Here, we propose a self-adaptive class weights evaluation mechanism. It includes two steps. First, reasonable and dynamic thresholds are set to cluster source classes into one, two and three groups respectively. Second,  Calinski-Harabaze index~\cite{in} is utilized to evaluate the optimal number of groups we cluster into.

If the optimal group number is three, the source classes are clustered into the shared, outlier and confused groups. It means that the model considers some classes (the confused group) are difficult to distinguish if they are from the shared label space or not. In this way, we can avoid misclassification due to the entanglement between shared and outlier domain and thus eliminate negative transfer greatly.

If the optimal group number is two, the source classes are clustered into the shared and outlier groups, which means the model can clearly distinguish all the source classes.  

Specially, if there is only one group, the shared group, the partial domain adaptation degrades to a standard domain adaptation.

After determining shared, outlier and confused classes, we can evaluate class weigth $w_i$. For samples of shared classes, $w_i$ is set to $1$. For samples of outlier classes, $w_i$ is set to $0$. For samples of confused classes, $w_i$ is set to $\alpha_{m(m=3)}$, which is the mean of confused classes' weight values in $W_c$ and will be specified later in the subsection.

The complete steps of $G_w$ are as follow:

\noindent
\textbf{Step 1. Determine optimal cluster given number of groups}

Define k as the number of groups ($1\le k<4$), and $S_m$ as the set of source classes belonging to the $m^{th}$ group $(1 \le m \le k)$. $\alpha_m$ is the mean of weight values in $W_c$ that belong to the set $S_m$. Illustration of these variables can be viewed in Fig.\ref{fig:varrela} (when k=3).

The intra-class variance $ssd(S_m)$ is obtained as below:
\begin{align}
ssd(S_m) &= \frac{1}{|C_s|}\sum_{j\in S_m}(W_c[j]-\alpha_m)^2 \\
\alpha_m &= \frac{\sum_{j\in S_m} W_c[j]}{|S_m|}
\end{align}

$\delta_k^2$ is the total of intra-class variance of the k groups.And it can be defined as:
\begin{align}
    \delta^2_{k}=\sum_{m=1}^{k}ssd(S_m)
\end{align}

Combining (5) and (6), we have:
\begin{align}
     \delta^2_{k}=\sum_{m=1}^k\sum_{j\in S_m}W_c^2[j]-\sum_{m=1}^k\frac{\sum_{j\in S_m}W_c^2[j]}{\sum_{j\in S_m}|S_m|}
\end{align}

By iteratively minimizing $\delta_{k}^2$, we can determine the best arrangement of the source classes in the $S_m$. The calculated optimal classification can determine the interruption point of the value in the $W_c$ by minimizing $ssd(S_m)$ within the $m^{th}$ group. A basic assumption is, if $\delta_{k}^2$ is minimal, the corresponding cluster situation is the best cluster result for $k$ groups cluster~\cite{Je}. Hence, we need to find out the minimal $\delta_{k}^2$ to effectively cluster the elements in $W_c$ under the circumstance of $k$ clusters.
Since $\sum_{m=1}^{k}\sum_{j\in S_m} W_c^2[j]$ is independent of $k$, minimizing $\delta_{k}^2$ is equivalent to :
\begin{align}
   \max \sum_{m=1}^k\frac{\sum_{j\in S_m}W_c^2[j]}{\sum_{j\in S_m}|S_m|}
\end{align}

Hence, when $\sum_{m=1}^k\frac{\sum_{j\in S_m}W_c^2[j]}{\sum_{j\in S_m}|S_m|}$ achieves its maximum, the optimal cluster results are obtained under the condition of k groups.

\noindent
\textbf{Step 2. Determine the optimal number of groups}

Now that we have got cluster results for different cluster number $k\in\{1,2,3\}$, the next step is to find out which cluster number is the best.

We utilize the Calinski-Harabasz index (CH index)~\cite{in} to compute the optimal $k$. The CH index depicts the tightness through the intra-class dispersion matrix and the separation through the inter-class dispersion matrix.
The inter-class dispersion matrix B(k) is given by:
\begin{align}
    B(k)=\sum_{m=1}^{|C_s|}\frac{|S_m|}{|C_s|}(\alpha_m-\frac{\sum_{j-1}^{|C_s|}W_c[j]}{|C_s|})(\alpha_m-\frac{\sum_{j-1}^{|C_s|}W_c[j]}{|C_s|})^T
\end{align}
We also set
\begin{align}
T_m=\frac{\sum_{j \in S_m} W_c[j]}{|S_m|}
\end{align}
The intra-class dispersion matrix S(k) is given by:
\begin{align}
    S(k)=\frac{1}{k}\sum_{m=1}^k\sum_{j \in S_m}(W_c[j]-T_m)(W_c[j]-T_m)^T
\end{align}
and the CH index is defined as
\begin{align}
    CH(k)=\frac{tr B(k)/(k-1)}{tr S(k)/(|C_s|-k)}
\end{align}
Here, $k$ is the current cluster group number. $trS(k)$ is the trace of the intra-class dispersion matrix, while $trB(k)$ represents the trace of the inter-class dispersion matrix.

Under the condition of $k$ groups, the larger CH($k$) is, the smaller the intra-group distance is, the greater inter-group distance is, the better the clustering result is.

Hence ,we get the follow relation:
\begin{align}
    k^*=\arg \max_{k}CH(k)
\end{align}

\noindent
\textbf{Step 3. Evaluate self-rectifying class weights $w_i$}

If the best group number $k^*=3$, source classes are clustered into three groups: shared classes $S_1$, outlier classes $S_2$ and confused classes $S_3$. For any source sample $(x_i^s,y_i^s)$, its weight $w_i$ is provided as:
\begin{equation}\label{wi}
w_i =\left\{
\begin{aligned}
& 1 \quad (x_i,y_i)\in S_1 \\
& 0 \quad (x_i,y_i)\in S_2 \\
& \alpha_{m(=3)} \quad (x_i,y_i)\in S_3
\end{aligned}\right.
\end{equation}
In this case, Our model not only maintains the discriminative ability to the identified shared samples and outlier ones, but also avoids misjudging the difficult-to-identify samples.
 
If $k^*=2$, it just includes two groups $S_1(D_{sh})$ and $S_2(D_{so})$. $w_i$ can be defined as:
\begin{equation}
w_i =\left\{
\begin{aligned}
                                  & 1 \quad (x_i,y_i)\in S_1 \\ & 0 \quad (x_i,y_i)\in S_2
\end{aligned}\right.
                                 \end{equation}
                                 
In this situation, the model believes it can distinguish the shared classes and outlier classes clearly, so the weights of shared samples are 1, and the weights of the outlier ones are 0. As a result, we can exclude the impact of the outlier classes as much as possible.

Specially, if best group number is  $k^*=1$, the model perceives the label space of source and target domains are the same. Partial domain adaptation degrades to standard domain adaptation. All the its weights $w_i$ are 1.
 
\subsection{Weighted Source Classifier and Domain Classifier}

\noindent
\textbf{Weighted Source Classifier}

The major challenge in PDA is that the class space of target domain is a subset of source ones~\cite{Jian2}, so source classifiers $G_c$ have poor results in target domain tasks due to the negative transfer caused by outlier classes.

Recall that $w_i$ represents the probability of sample $x_i$ belonging to shared classes. Using $w_i$ as weight values, we can solve the above problem by paying less attention to the outlier classes and only focus on the shared classes. On the other hand, we use an extra weight $m_i$ to draw together samples from the same class in ${\mathcal D}_{sh}$ and ${\mathcal D}_t$ and push away outlier samples. In summary, we propose weighted source classifier whose loss is defined as follows:
\begin{equation}
\begin{split}
    L_{ad}(G_c)=&\frac { 1 } { n _ { s } } \sum _ { \mathbf i=1 }^{n_s}   w_i(x_i^s) m_i(x_i^s)  L  \left( G_c\left( G_f  \left( x_{i}^s \right) \right) , y _ { i } \right)
\end{split}
\end{equation}

where
\begin{equation}
\begin{split}
    m(x) &= 1 + e^{-H(x)} \\
    H\left ( x \right ) &= -\sum p\left (  x \right )log\left (p\left (  x \right )  \right ) \\
    p(x) &= G_c(G_f(x))
\end{split}
\end{equation}

Here, the information entropy loss is employed in both domains so that they are harder to be misclassified. Because the target samples are unlabeled, they are much easier to tangle with each other. It is worth noting that at first $w_i$ is set as 1, it will update each a few iterations regularly.

\noindent
\textbf{Weighted Domain Adaptation Framework}

In subsection B, we introduce a paradigm involving $G_d$ and $G_f$ that learns domain-invariant features. Though this paradigm is effective, negative transfer will arise if we apply it to partial domain adaptation directly. Because there is still incongruity between source and target label spaces.

Therefore, we similarly use weight values $w_i$ and $m_i$ to enhance positive transfer and alleviate negative transfer. Specifically, weights for samples in ${\mathcal D}_{sh}$ are promoted and weights for samples in ${\mathcal D}_{so}$ are decreased. The weighted adversarial domain adaptation framework for PDA can be defined as follows:
\begin{equation}
\begin{split}
&\min _ { G_f } \max _ { G_d }{L_d}( G_f,G_d ) \\
&= \mathbb { E } _ { x  \sim p _ { s } } \left[w_{i}(x_i) m_{i}(x_i) log \left( G_d ( G_f\left ( x_i \right ) \right)) \right]\\ &+ \mathbb { E } _ { x \sim p _ { t } } \left[  log\left(1- G_d \left( G_f ( x_i \right ) \right)) \right] 
\end{split}
\end{equation}

The higher $w_i$ is, the more likely the sample is from shared classes.

\subsection{Cluster Classifier}
As mentioned in subsection B, $G_{cl}$ classify samples into shared, oulier and confused clusters. $G_{cl}$ helps the model recognize shared domain samples by separating each cluster.

The problem is that cluster labels is not originally available like class labels. As a solution, we use $w_i$ as cluster labels. The reason is that, as indicated by Eq.\ref{wi} and Fig.\ref{fig:varrela}, samples from the same cluster $S_m$ have the same weight value $w_i$. So it is reasonable to use $w_i$ as cluster labels. Specially, cluster labels for target domain samples are 1 because they ought to be clustered with share domain samples.

By utilizing the cross-entropy loss function as $L$, the loss function $L_{cl}$  of $G_{cl}$ is:
\begin{equation}\label{19}
\begin{split}
L_{cl}(G_{cl})=&\frac{1}{n_s}\sum_{i=1}^{n_s}{L(G_{cl}(G_f(x_i^s),w_i))}\\&+\frac{1}{n_t}\sum_{i=1}^{n_t}{L(G_{cl}(G_f(x_i^t),1))}
\end{split}
\end{equation}

By minimizing $L_{cl}$, the model reduces the distance among samples from the same cluster, and extends the distance among samples from different clusters.

During training, if $k^*=2$, $L_{cl}$ can be set as follows:
\begin{equation}\label{20}
\begin{split}
L_{cl}(G_{cl})=&\frac{1}{n_{s}}(\sum_{i=1}^{n_{sh}}{L(G_{cl}(G_f(x_i^{sh}),1))}\\&+\sum_{i=1}^{n_{ou}}{L(G_{cl}(G_f(x_i^{so}),0))})\\&+\frac{1}{n_t}\sum_{i=1}^{n_t}{L(G_{cl}(G_f(x_i^t),1))}
\end{split}
\end{equation}

In Eq.\ref{20}, $n_{sh}$ is denoted as the number of samples from shared domain, and $n_{ou}$ as the number of samples from outlier domain. We have $n_{sh}+ n_{ou}=n_s$.

By this means, source shared and target samples can be clustered together. At the same time, outlier ones can be separated away, which decreases negative transfer greatly.

Moreover, if our SAPDA regards the situation as a standard domain adaptation setting, source and target domain share the same cluster label 1. In this situation,  $L_{cl}$ can be written as follow:
\begin{equation}
\begin{split}
L_{cl}(G_{cl})&=\frac{1}{n_{s}}(\sum_{i=1}^{n_{s}}{L(G_{cl}(G_f(x_i^{s}),1))}\\&+\frac{1}{n_t}\sum_{i=1}^{n_t}{L(G_{cl}(G_f(x_i^t),1))}
\end{split}
\end{equation}

\subsection{Self-adaptive Partial Domain Adaptation }
A novel self-adaptive partial domain adaptation framework is proposed to handle PDA task. This framework can self-adaptively cluster source domain into different groups to progressively measure the transferability of source classes on sample level by weighting samples in the same group equally, and jointly learn domain-invariant features across different domains. The complete algorithm is as follow:

\begin{algorithm}
\caption{SAPDA}
\begin{algorithmic}
\State \textbf{Input}: labeled source data $\{(x_i^s,y^s_i)\}_{i=1}^{n_s}$ and unlabeled target data $\{{x}_i^t\} _{i = 1}^{{n_t}}$
\State \textbf{Output}: predicted labels $\{{y}_i^t\}_{i=1}^{{n_t}}$
\State \textbf{Initialization}: $w_i$ $\leftarrow$ 1 for all samples
\For {$iteration(i)=1,2,\ldots, 16000$}
    \State 1) Extract features from one batch: 
    \State \quad$(f_1^s,...,f_{batch}^s,f_1^t,...,f_{batch}^t)$ 
    \State$ \quad= G_f(x_1^s,...,x_{batch}^s,x_1^t,...,x_{batch}^t)$
    \State 2) Classify each samples in one batch and use eq.(16)
    \State \quad to calculate loss:
    \State \quad$(g_1^s,...,g_{batch}^s,g_1^t,...,g_{batch}^t)$
    \State$\quad = G_c(f_1^s,...,f_{batch}^s,f_1^t,...,f_{batch}^t)$
    \State 3) Uses eq.(18) to calculate loss for classifier $G_d$
    \State 4) Uses eq.(19)/(20)/(21) to calculate loss for $G_{cl}$
    \State 5) Loss back propagation
    \If {$i \,\%\, 500 == 0$}
	\State a) Update $W_c$ using $W_c=\frac{1}{n_t}\sum_{i=1}^{n_t}(G_c(G_f(x_i^t)))$.
	\State b) Update $w_i$ using eq.(14)/(15) or $w_i$ $\leftarrow$ 1 for all \State \quad samples
	\EndIf
\EndFor
\end{algorithmic}
\end{algorithm}

$\theta_{c}$, $\theta_{cl}$, $\theta_{f}$ and $\theta_{d}$ are respectively the parameters of $G_c$, $G_{cl}$, $G_f$ and $G_d$. A saddle point solution ($\hat{\theta_{c}}$, $\hat{\theta_{cl}}$, $\hat{\theta_{f}}$, $\hat{\theta_{d}}$) is achieved by a end-to-end minimax optimization procedure:
\begin{equation}
\begin{split}
   &(\hat{\theta}_{c})=\arg \min_{\theta_{c}} L_c(G_c)
  \\&(\hat{\theta}_{cl})=\arg  \min_{\theta_{cl}}L_{cl}(G_{cl})
  \\&(\hat{\theta}_{c}, \hat{\theta}_{cl}) = \arg \min_{\theta_{c}, \theta_{cl}} (L_c(G_c) + \beta L_{cl}(G_{cl}))
\end{split}
\end{equation}


\begin{table*}[tb!]
    \large
    \centering 
    \caption{Accuracy of partial DA tasks on \emph{Caltech-Office} (10 classes $\rightarrow$ 5 classes).}
    \label{table:accuracy_co1}
        \vspace{-10pt}
  \resizebox{0.95\textwidth}{!}{%
  \renewcommand\tabcolsep{2.0pt}
    \begin{tabular}{r|cccccccccccc|c}
        \hline
        \multirow{2}{45pt}{\centering Method} &
        \multicolumn{13}{c}{Caltech-Office (10 classes $\rightarrow$ 5 classes)}
        \\
        \cline{2-14}
        & C $\rightarrow$ A & C $\rightarrow$ W & C $\rightarrow$ D & A $\rightarrow$ C & A $\rightarrow$ W & A $\rightarrow$ D & W
        $\rightarrow$ C & W
        $\rightarrow$ A & W $\rightarrow$ D & D $\rightarrow$ C & D $\rightarrow$ A & D $\rightarrow$ W & Avg \\
        \hline
        AlexNet~\cite{Krizhevsky}& 93.58 & 83.70 & 91.18 & 85.27 & 76.30 & 85.29 & 74.17
        &87.37 & \textbf{100.00} & 80.82 & 89.51 & 98.52 & 98.52\\
        DaNN~\cite{Ganin16}& 91.86 & 82.22 & 83.82 & 77.57 & 65.93 & 80.88 & 72.60
        &80.30 & 95.59 & 69.35 & 77.09 & 80.74 &79.83 \\
        RTN~\cite{Long17}& 91.86 & 93.33 & 80.88 & 80.99 & 69.63 & 70.59 & 59.08 
        & 74.73 & \textbf{100.00} & 59.08 & 70.02 & 91.11 & 78.44 \\
        ADDA~\cite{Tzeng17}& 93.15 & 94.07 & 97.06 & 85.27 & 87.41 & 89.71 & 86.82 & 92.08 &\textbf{100.00} & 89.90 & 93.79 & 98.52 & 92.31\\
        IWAN~\cite{Zhang18}& 94.22 & 97.78 & 98.53 & 89.90 & 87.41 & 88.24 & 90.24 & 95.29 & 
        \textbf{100.00} & 91.61 & 94.43 & 98.52 & 93.85\\
        PADA~\cite{Cao18}& 96.25 & 96.00 & 97.59 & 92.05 & 87.33 & 96.39 & 96.85 & 96.14
        &\textbf{100.00} & 95.80 & 97.31 & 97.87 & 95.72\\
        ETN ~\cite{Cao19}& 96.16 & 96.02 & 98.33 & 95.13 & 90.01 & 98.54 & 96.06 & 96.66
        &\textbf{100.00} & 96.00 & 96.14 & 97.93 & 96.42\\
        \hline
          SAPDA & \textbf{97.11} & \textbf{98.98} & \textbf{99.07} & \textbf{97.60} & \textbf{92.00} & \textbf{100.00} & \textbf{97.80} &
        \textbf{97.52} &
        \textbf{100.00} &
        \textbf{98.05} &
        \textbf{96.90} &
        \textbf{100.00} &
        \textbf{97.92}
        \\ 
        \hline
    \end{tabular}%
}
\end{table*}

\begin{table*}[tb!]
    \large
    \addtolength{\tabcolsep}{6pt}
    \centering 
    \caption{Accuracy of partial domain adaptation tasks on \emph{Office-31}}
    \label{table:accuracy_officeic}
    \scalebox{0.8}{%
    \begin{tabular}{c|cccccc|c}
        \hline
        \multirow{2}{42pt}{\centering Method} &  \multicolumn{7}{c}{Office-31} \\
        \cline{2-8}
        & A $\rightarrow$ W & D $\rightarrow$ W & W $\rightarrow$ D & A $\rightarrow$ D & D $\rightarrow$ A & W $\rightarrow$ A & Avg \\
        \hline
        ResNet ~\cite{He}& 54.52 & 94.57 & 94.27 & 65.61 & 73.17 & 71.71 & 75.64\\
        DAN ~\cite{Long151}& 46.44 & 53.56 & 58.60 & 42.68 & 65.66 & 65.34 & 55.38 \\
        DaNN~\cite{Ganin16}& 41.35 & 46.78 & 38.85 & 41.36 & 41.34 & 44.68 & 42.39 \\
        RTN ~\cite{Long17}& 75.25 & 97.12 & 98.32 & 66.88 & 85.59 & 85.70 & 84.81 \\
        IWAN ~\cite{Zhang18}& 76.27 & 98.98 & \textbf{100.00} & 78.98 & 89.46 & 81.73 & 87.57 \\
        SAN ~\cite{Cao181}& 81.82 & 98.64 & \textbf{100.00} & 81.28 & 80.58 & 83.09 & 87.27 \\
        PADA ~\cite{Cao18}& 86.54 & 99.32 & \textbf{100.00} & 82.27 & 92.69 & 95.41 & 92.69 \\ 
         ETN ~\cite{Cao19}& 94.52 & \textbf{100.00} & \textbf{100.00} & 95.03 & 96.21 & 94.54 & 96.73 \\ 
        \hline
        SAPDA & \textbf{96.61} & \textbf{100.00} & \textbf{100.00} & \textbf{97.45} & \textbf{96.49} & \textbf{95.83} & \textbf{97.73} \\ 
        \hline
    \end{tabular}%
}
\end{table*}

\begin{table*}[tb!]
    \large
    \centering 
    \caption{Accuracy of partial domain adaptation tasks on \emph{VisDA2017}(12 classes $\rightarrow$ 6 classes) and \emph{Caltech-Office}(256 classes $\rightarrow$ 10 classes)}
    \label{table:accuracy_office}
    \scalebox{0.8}{%
    \begin{tabular}{r|c|c|ccc|c}
        \hline
        \multirow{2}{40pt}{\centering Method} &  \multicolumn{1}{c|}{\centering VisDA2017} &
        \multirow{2}{*}{\centering Method} &  \multicolumn{4}{c}{\centering Caltech-Office}\\
        \cline{2-2}
        \cline{4-7}
        & S $\rightarrow$ R & &
        C $\rightarrow$ W & C $\rightarrow$ A & C $\rightarrow$ D & Avg\\
        \hline
        ResNet ~\cite{He}& 45.62 & AlexNet ~\cite{Krizhevsky}& 58.44 & 74.64 & 65.86 & 66.98\\
        DaNN ~\cite{Ganin16}& 51.01 & ResNet ~\cite{He}& 61.33 & 77.57 & 68.90 & 69.27 \\
        RTN ~\cite{Long17}& 50.04 & DaNN~\cite{Ganin16}& 54.57 &72.86 & 57.96 & 61.80 \\
        IWAN ~\cite{Zhang18}& 52.18 &RTN ~\cite{Long17}& 71.02 & 81.32 & 62.35 & 71.56\\
        SAN ~\cite{Cao181}& 52.06  & DAN ~\cite{Long151}& 42.37 & 70,75 & 47.04 & 53.39 \\
        PADA ~\cite{Cao18}& 53.53 & SAN ~\cite{Cao181}& 88.33 & 83.87 & 85.54 & 85.83 \\ 
         ETN ~\cite{Cao19}& 57.09 & PADA ~\cite{Cao18}& 89.07 & 89.34 & 88.54 & 88.93\\ 
        \hline
        SAPDA & \textbf{59.87} & SAPDA & \textbf{89.83} & \textbf{92.93} & \textbf{90.45} & \textbf{91.07}  \\ 
        \hline
    \end{tabular}%
}
\end{table*}

\begin{table*}[tb!]
    \large
    \centering 
    \caption{Accuracy of partial DA tasks on \emph{Office-Home} (65 classes $\rightarrow$ 25 classes).}
    \label{table:accuracy_officehome}
        \vspace{-10pt}
  \resizebox{1.0\textwidth}{!}{%
  \renewcommand\tabcolsep{1.0pt}
    \begin{tabular}{r|cccccccccccc|c}
        \hline
        \multirow{2}{45pt}{\centering Method} &
        \multicolumn{13}{c}{Office-Home}
        \\
        \cline{2-14}
        & Ar $\rightarrow$ Cl & Ar $\rightarrow$ Pr & Ar $\rightarrow$ Rw & Cl $\rightarrow$ Ar & Cl $\rightarrow$ Pr & Cl $\rightarrow$ Rw & Pr
        $\rightarrow$ Ar & Pr
        $\rightarrow$ Cl & Pr $\rightarrow$ Rw & Rw $\rightarrow$ Ar & Rw $\rightarrow$ Cl & Rw $\rightarrow$ Pr & Avg \\
        \hline
        ResNet-50~\cite{He}& 46.33 & 67.51 & 75.87 & 59.14 & 59.94 & 62.73 & 58.22 & 41.79 & 74.88 & 67.40 & 48.18 & 74.17 & 61.35\\
        DaNN~\cite{Ganin16}& 43.76 & 67.90 & 77.47 & 63.73 & 58.99 &  67.59 & 56.84 & 37.07 & 76.37 &  69.15 & 44.30 & 77.48 & 61.72 \\
        RTN~\cite{Long17}& 49.31 & 57.70 & 80.07 & 63.54 & 63.47 & 73.38 & 65.11 & 41.73 & 75.32 & 63.18 & 43.57 & 80.50 & 63.07 \\
        IWAN~\cite{Zhang18}& 53.94 & 54.45 & 78.12 & 61.31 & 47.95 & 63.32 & 54.17 & 52.02 & 81.28 & 76.46 & 56.75 & 82.90 & 63.56\\
        SAN~\cite{Cao181}& 44.42 & 68.68 & 74.60 & 67.49 & 64.99 & 77.80 & 59.78 & 44.72 & 80.07 & 72.18 & 50.21 & 78.66 & 65.30\\
        PADA~\cite{Cao18}& 51.95 & 67.00 & 78.74 & 52.16 & 53.78 & 59.03 & 52.61 & 43.22 & 78.79 & 73.73 & 56.60 & 77.09 & 62.06\\
        MWPDA~\cite{Jian1}& 55.39 & 77.53 & 81.27 & 57.08 & 61.03 & 62.33 & 68.74 & 56.42 & 86.67 & 76.70 & 57.67 & 80.06 & 68.41\\
        ETN ~\cite{Cao19}& 59.24 & 77.03 & 79.54 & 62.92 & 65.73 & 75.01 & 68.29 & 55.37 & 84.37 & 75.72 & 57.66 & 84.54 & 70.45
        \\
         B$A^3$US~\cite{liang2020balanced}& 60.62 & \textbf{83.16} & \textbf{88.39} & 71.75 & 72.79 & \textbf{83.40} & 75.45 & 61.59 & 86.53 & 79.25 & 62.80 & 86.05 & 75.98
        \\
        \hline
          SAPDA & \textbf{63.81} & 82.55 & 85.66 & \textbf{72.34} & \textbf{73.07} & 82.66 & \textbf{77.64} &
        \textbf{62.90} &
        \textbf{88.64} &
        \textbf{80.15} &
        \textbf{63.55} &
        \textbf{86.29} &
        \textbf{76.61}
        \\ 
        \hline
    \end{tabular}%
}
\end{table*}

\section{Experiment}
To illustrate the performance of SAPDA, we conduct some experiments on four benchmark compared with previous standard and partial domain adaptation methods.

\subsection{Set up}
\par \textbf{Office-31}~\cite{Saenko10} dataset is a classic dataset for domain adaptation. It involves three domains: DSLR, Amazon, and Webcam, we denote them as D31, A31 and W31 respectively. They are set as source domains. There are 10 categories~\cite{Gong} shared by Caltech-256~\cite{Griffin} and Office-31 dataset. These 10 categories, denoted as W10, A10 and D10, are set as target domain. Moreover, Caltech-256 are also set as source domain and A10, W10 and D10 are set as target domain on three tasks.
\begin{table*}[tb!]
    \large
    \addtolength{\tabcolsep}{6pt}
    \centering 
    \caption{Accuracy on Partial DA tasks of SAPDA and its variants on \emph{Office-31}(31 classes $\rightarrow$ 10 classes)}
    \label{table:accuracy_officeviance}
    \scalebox{0.8}{%
    \renewcommand\tabcolsep{3.0pt}
    \begin{tabular}{c|cccccc|c}
        \hline
        \multirow{2}{30pt}{\centering Method} &  \multicolumn{7}{c}{Office-31} \\
        \cline{2-8}
        & A $\rightarrow$ W & D $\rightarrow$ W & W $\rightarrow$ D & A $\rightarrow$ D & D $\rightarrow$ A & W $\rightarrow$ A & Avg \\
        \hline
        SAPDA w/o self-adaptive class weights evaluation mechanism & 87.09 & \textbf{100.00} & \textbf{100.00} & 81.13 & 91.26
        & 95.30 & 92.46  \\
        
        SAPDA w/o cluster classifier & 91.07 & \textbf{100.00} & \textbf{100.00} & 96.45 & 95.02
        & 95.42 & 96.33 \\
        
        SAPDA with shared, outlier and confused classes & 92.03 & 97.53 & 98.56 & 94.37 & 92.06
        & 94.02 & 94.76  \\
        
        SAPDA with shared and outlier classes & 95.37 & \textbf{100.00} & \textbf{100.00} & 95.01 & 95.02 & 95.39 & 96.79 \\
        
        SAPDA & \textbf{96.61} & \textbf{100.00} & \textbf{100.00} & \textbf{97.45} & \textbf{96.49} & \textbf{95.83} & \textbf{97.73} \\
        \hline
    \end{tabular}%
    
}

\end{table*}
\begin{table*}[tb!]
    \large
    \centering 
    \caption{Accuracy of standard DA tasks on \emph{Office-Home} (65 classes $\rightarrow$ 65 classes).}
    \label{table:accuracy_std}
        \vspace{-10pt}
  \resizebox{1.0\textwidth}{!}{%
  \renewcommand\tabcolsep{2.0pt}
    \begin{tabular}{r|cccccccccccc|c}
        \hline
        \multirow{2}{45pt}{\centering Method} &
        \multicolumn{13}{c}{Office-Home}
        \\
        \cline{2-14}
        & Ar $\rightarrow$ Cl & Ar $\rightarrow$ Pr & Ar $\rightarrow$ Rw & Cl $\rightarrow$ Ar & Cl $\rightarrow$ Pr & Cl $\rightarrow$ Rw & Pr
        $\rightarrow$ Ar & Pr
        $\rightarrow$ Cl & Pr $\rightarrow$ Rw & Rw $\rightarrow$ Ar & Rw $\rightarrow$ Cl & Rw $\rightarrow$ Pr & Avg \\
        \hline
        ResNet-50~\cite{He}& 34.9 & 50.0 & 58.0 & 37.4 & 41.9 & 46.2 & 38.5 & 31.2 & 60.4 & 53.9 & 41.2 & 59.9 & 46.1\\
        DaNN~\cite{Ganin16}& 45.6 & 59.3 & 70.1 & 47.0 & 58.5 & 60.9 & 46.1 & 43.7 & 68.5 & 63.2 & 51.8 & 76.8 & 57.6 \\
        JAN~\cite{Tzeng17}& 45.9 & 61.2 & 68.9 & 50.4 & 59.7 & 61.0 & 45.8 & 43.4 & 70.3 & 63.9 & 52.4 & 76.8 & 58.3 \\
        DAN~\cite{Zhang18}& 45.6 & 67.7 & 73.9 & 57.7 & 63.8 & 66.0 & 54.9 & 40.0 & 74.5 & 66.2 & 49.1 & 77.9 & 61.4\\
        CDAN+E~\cite{2017Conditional}& 50.7 & 70.6 & 76.0 & 57.6 & 70.0 & 70.0 & 57.4 & 50.9 & 77.3 & 70.9 & 56.7 & 81.6 & 65.8\\
        DRCN~\cite{li2020deep}& 50.6 & 72.4 & 76.8 & \textbf{61.9} & 69.5 & 71.3 & \textbf{60.4} & 48.6 & 76.8 & \textbf{72.9} & 56.1 & 81.4 & 66.6\\
        \hline
          SAPDA & \textbf{51.5} & \textbf{71.3} & \textbf{76.9} & 59.0 & \textbf{71.3} & \textbf{71.6} & 58.6 &
        \textbf{51.2} &
        \textbf{77.6} & 71.5 &
        \textbf{59.1} &
        \textbf{82.6} &
        \textbf{66.9}
        \\ 
        \hline
    \end{tabular}%
}
\end{table*}

\par \textbf{Caltech-Office} dataset utilizes 10 classes shared by Caltech-256 and Office-31 as source domain, denoted as W10, D10, A10 and C10, the first 5 classes in these 10 categories, denoted as W5, D5, A5 and C5, are set as target domain. 

\par \textbf{Office-Home} dataset~\cite{Venkateswara} is a much more intriguing dataset with the huger domain gap. It includes 65 categories with four domains: Artistic, Product images, Real-World and Clip Art. We define 65 classes in four domains as source domain Ar, Pr, Rw and Cl. Four domains with the first 25 classes are donated as target domain.

\par \textbf{VisDA-2017} dataset is one of the most challenging dataset in domain adaptation, the synthetic data to real-image track is evaluated here. Under our partial domain adaptation setting, the first 6 classes are chosen as target domain and Synthetic 12 $\rightarrow$ Real6 task is conducted as S $\rightarrow$ R. 

\par The proposed SAPDA is compared with present standard DA and PDA methods. Among all the experiments, both standard DA and PDA methods are perform on PDA setting. ResNet-50 are used as the base backbone for all the methods except for AlexNet. Meanwhile, classic supervised learning methods like ResNet-50 train on the labeled source domain and test on the unlabeled target domain. 
\par Furthermore, plenty of ablation experiments are carried on by assessing four variants of SAPDA: 1) SAPDA w/o self-adaptive class weights evaluation mechanism is the variant without self-adaptive class weights evaluation, degenerating to PADA with cluster classifier. 2) SAPDA w/o cluster classifier is the variant without cluster classifier. 3) SAPDA with shared, outlier and confused classes is the variant takes the weights with shared, outlier and confused classes. 4) SAPDA with shared and outlier classes is the variant that takes the weights with only shared and outlier classes.

\par Our implementation is based on PyTorch, and finetune pre-trained ResNet-50~\cite{He}. Similar to DaNN, a bottleneck layer is added after the feature extractor. Bottleneck layer, outlier discriminator $G_{cl}$ domain discriminator $G_d$ and feature extractor $G_f$ are trained from square one. Mini-batch stochastic gradient descent are used during the training.  We also select the same learning rate as DaNN. The learning rate is coordinated during training following  
$p =\frac{\gamma_{0}}{(1+\eta p)^\alpha}$, where $\eta$ and $\alpha$ are changed with importance-weighted cross-validation ~\cite{Sugiyama}, and p is the hyper-parameter optimized based on the dataset.

\subsection{Result}
The classification results on Office10-Caltech5, a set of Office-31 and VisDA-2017, a set of Caltech256-Office10 and Office-Home are respectively shown in Table~\ref{table:accuracy_co1}\textasciitilde~\ref{table:accuracy_officehome}. We also perform some ablation experiments in Table~\ref{table:accuracy_officeviance}. The results indicate our SAPDA outperforms all the standard DA and PDA methods.

We also have some insightful observations. \textbf{(1)} supervised methods like AlexNet and ResNet perform better on standard DA method under PDA setting, it shows negative transfer has negative impact on the accuracy when the features from outlier source classes are learned by standard DA methods such as DaNN and DAN. \textbf{(2)} RTN utilizes the entropy minimization criterion to modify the problem. Hence, it is an improvement over ResNet, but there is still some negative transfer on most tasks. \textbf{(3)} Since the weight mechanism can select the shared classes and promote their weights, PDA methods achieve better result than ResNet-50 and other standard DA methods. \textbf{(4)} Our SAPDA outperforms both standard DA and PDA methods, demonstrating that our self-adaptive weight mechanism can effectively utilize the confused class to avoid misjudging confused samples.
\begin{figure*}[tb!]
\centering
\subfigure[ResNet]{
\begin{minipage}[t]{0.248\linewidth}
\centering
\includegraphics[width=1.83in]{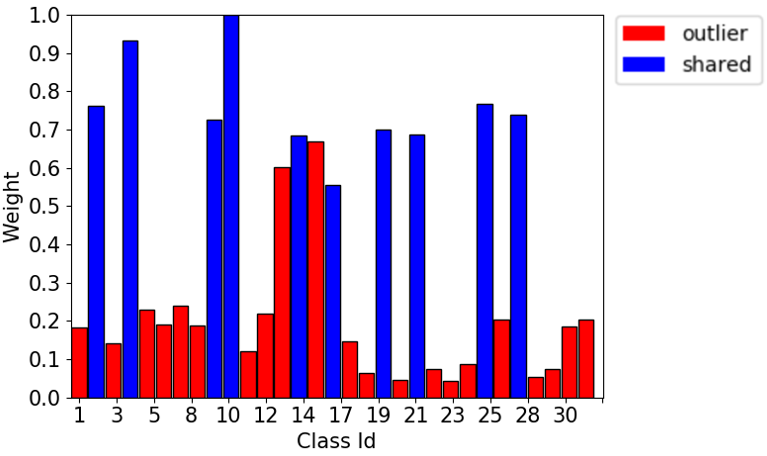}
\end{minipage}%
}%
\subfigure[DaNN]{
\begin{minipage}[t]{0.248\linewidth}
\centering
\includegraphics[width=1.83in]{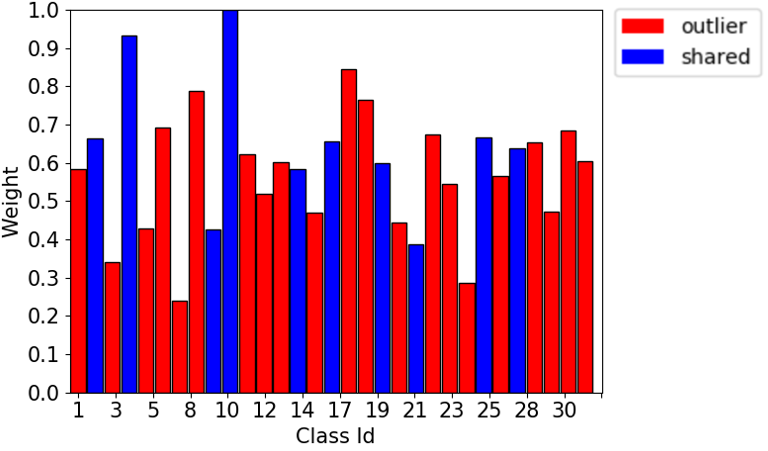}
\end{minipage}%
}%
\subfigure[PADA]{
\begin{minipage}[t]{0.24\linewidth}
\centering
\includegraphics[width=1.83in]{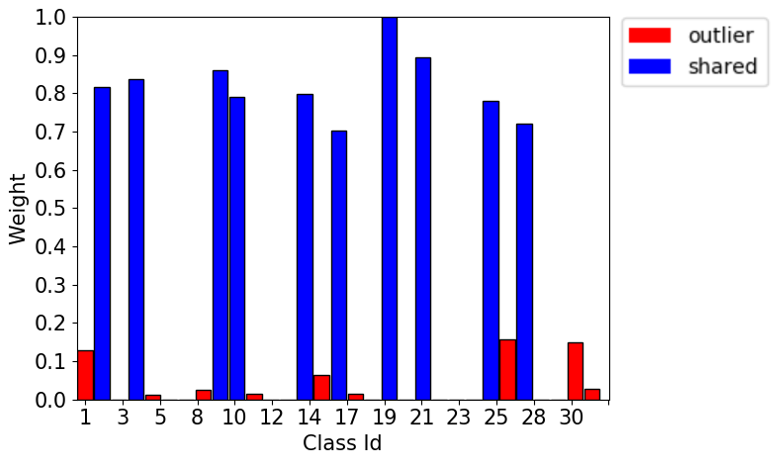}
\end{minipage}
}%
\subfigure[SAPDA]{
\begin{minipage}[t]{0.24\linewidth}
\centering
\includegraphics[width=1.83in]{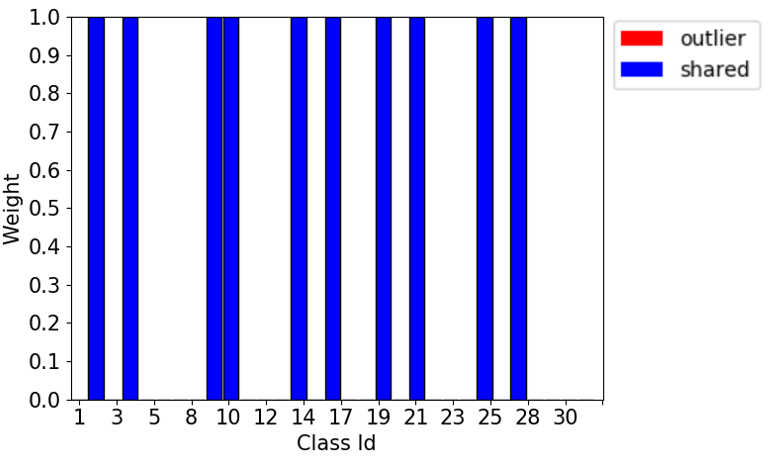}
\end{minipage}
}%

\caption{Comparison between class weights histograms learned from ResNet-50,DaNN, PADA and SAPDA}
\label{fig:weights}
\end{figure*}

\begin{figure*}[tb!]
\centering
\subfigure[ResNet]{
\begin{minipage}[t]{0.24\linewidth}
\centering
\includegraphics[width=1.7in]{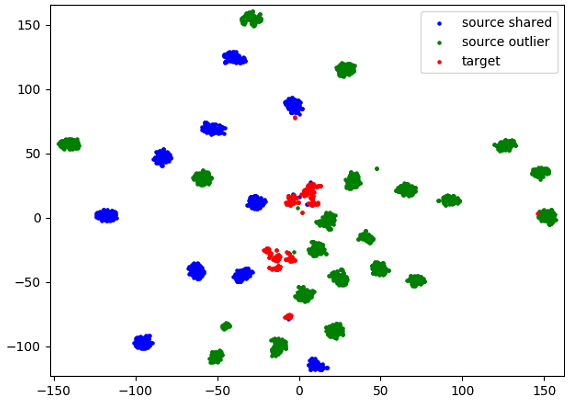}
\end{minipage}%
}%
\subfigure[DaNN]{
\begin{minipage}[t]{0.24\linewidth}
\centering
\includegraphics[width=1.7in]{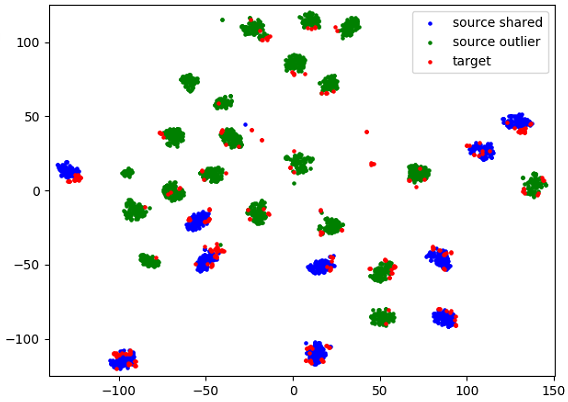}
\end{minipage}%
}%
\subfigure[PADA]{
\begin{minipage}[t]{0.24\linewidth}
\centering
\includegraphics[width=1.7in]{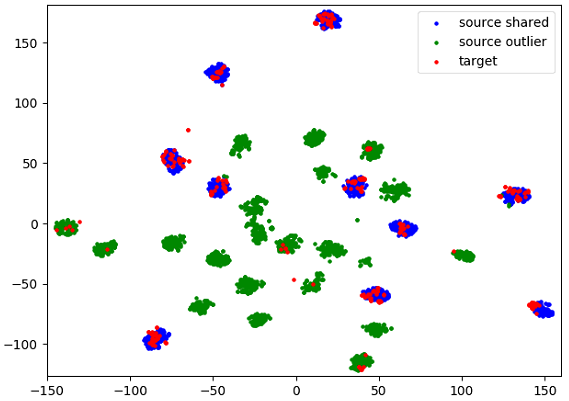}
\end{minipage}
}%
\subfigure[SAPDA]{
\begin{minipage}[t]{0.24\linewidth}
\centering
\includegraphics[width=1.7in]{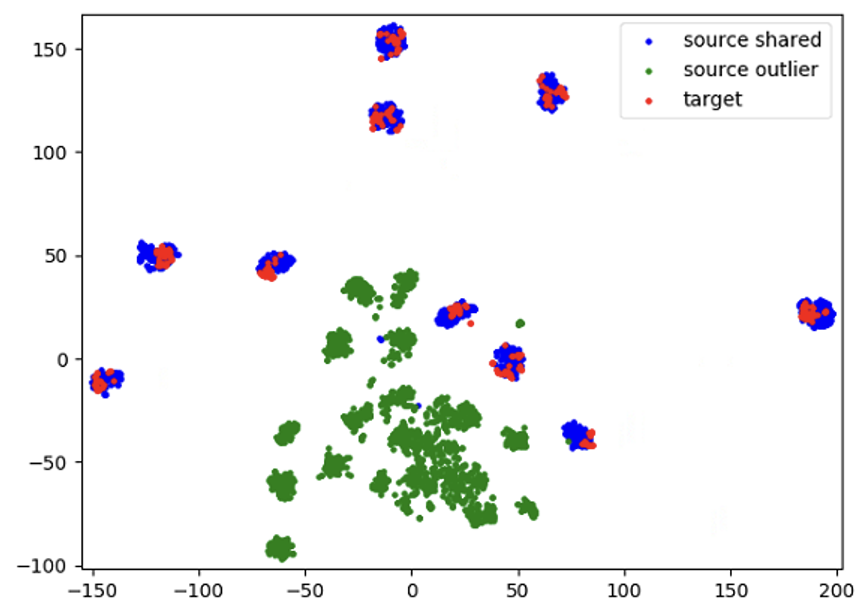}
\end{minipage}
}%
\caption{Visualization of features learned by ResNet,DaNN, PADA and SAPDA}
\label{fig:t-sne}
\end{figure*}

We further discover different components of SAPDA by contrast with the results of SAPDA variants in Tables~\ref{table:accuracy_officeviance}. \textbf{(1)} SAPDA outperforms SAPDA w/o self-adaptive class weights evaluation mechanism, proving that using self-adaptive class weights evaluation mechanism can select out reasonable outlier classes, further weaken the negative impact of outlier data, and force the source classifier to pay attention to data pertaining to the target label space. \textbf{(2)} SAPDA outperforms SAPDA w/o cluster classifier, showing the cluster classifier can gather different classes more tightly to avoid misclassification partly. \textbf{(3)} SAPDA with shared, outlier and confused classes gets the worst results on almost each task especially on the tasks D31 $\rightarrow$ W10 and W31 $\rightarrow$ D10. On these two tasks, because of the gap between different classes is small, it is easy for model to select out shared and outlier classes. But when we cluster source classes into shared, outlier and confused 
classes, neither the shared classes can be weighted as 1, nor the outlier classes are weighted as 0. This phenomenon can cause negative transfer, reducing accuracy, which also illustrate the necessity of putting $W_c$ to the expected values. Meanwhile, the result on task S $\rightarrow$ R performs much better than SAPDA with shared, outlier and confused classes. This shows that in a more challenging task, arbitrarily converting the soft $W_c$ to a binary value can easily cause the difficult-to-discern shared classes to be misjudged as outlier classes or vice versa. This result also illustrates the necessity of self cluster weights mechanism. \textbf{(4)} SAPDA with shared and outlier classes achieves close result to SAPDA on Office-31 dataset, but does not perform well on more difficult VisDA-2017. It indicates the self cluster weights mechanism can play a much more important role when the task is much more harder.

Moreover, we also apply our method to the standard domain adaptation problem, which is shown in Table~\ref{table:accuracy_std}, because our sample weight evaluation mechanism became a transferability degree measurement mechanism, which also helped to improve the classification accuracy under this setting.
\begin{table}[tb!]
\label{table:1}
    \large
    \centering 
    \caption{Accuracy on \emph{Office-31} by different $\beta$ of partial domain adaptation tasks}
    \label{table:accuracy_beta}
    \scalebox{0.85}{%
    \begin{tabular}{cccc}
        \hline
        {\centering hyper $\beta$}
        & A $\rightarrow$ W & A $\rightarrow$ D & Avg on Office-31\\
        \hline
        0.01& 95.77 & 96.90 & 96.54\\
        0.02& 95.74 & 97.30 & 96.22\\
        0.05& 95.81 & 97.22 & 96.98\\
        0.1& \textbf{96.61} & \textbf{97.45} & \textbf{97.73}\\
        0.5& \textbf{96.61} & 96.90 & 97.02\\
        1 & 95.77 & 97.22& 96.87\\
        \hline
    \end{tabular}%
}
\end{table}
\subsection{Analysis}
\textbf{Class Weight}: Fig.~\ref{fig:weights}(a)-(d) shows class weight histograms for task A (31 classes) $\rightarrow$ W (10 classes). They are learned by fineturing ResNet-50, DaNN, PADA and our SAPDA respectively. The blue bins are weights for shared classes and the red are the weights for outlier classes.

Fig.~\ref{fig:weights}(a) shows ResNet can distinguish some target samples due to finetune. Fig.~\ref{fig:weights}(b) implies negative transfer causes network cannot distinguish shared and outlier classes. Fig.~\ref{fig:weights}(c) illustrates although PADA can filter out shared classes, the weights of shared cannot achieve 1 while the weights of the outlier cannot achieve 0, which can still bring about performance reduction. Fig.~\ref{fig:weights}(d) shows our SPADA can not only select shared classes correctly, but can also promote the weights of shared to 1 and down else to 0, which can eliminate negative transfer greatly.  
\begin{figure}[tb]
\centering
\includegraphics[scale=0.4]{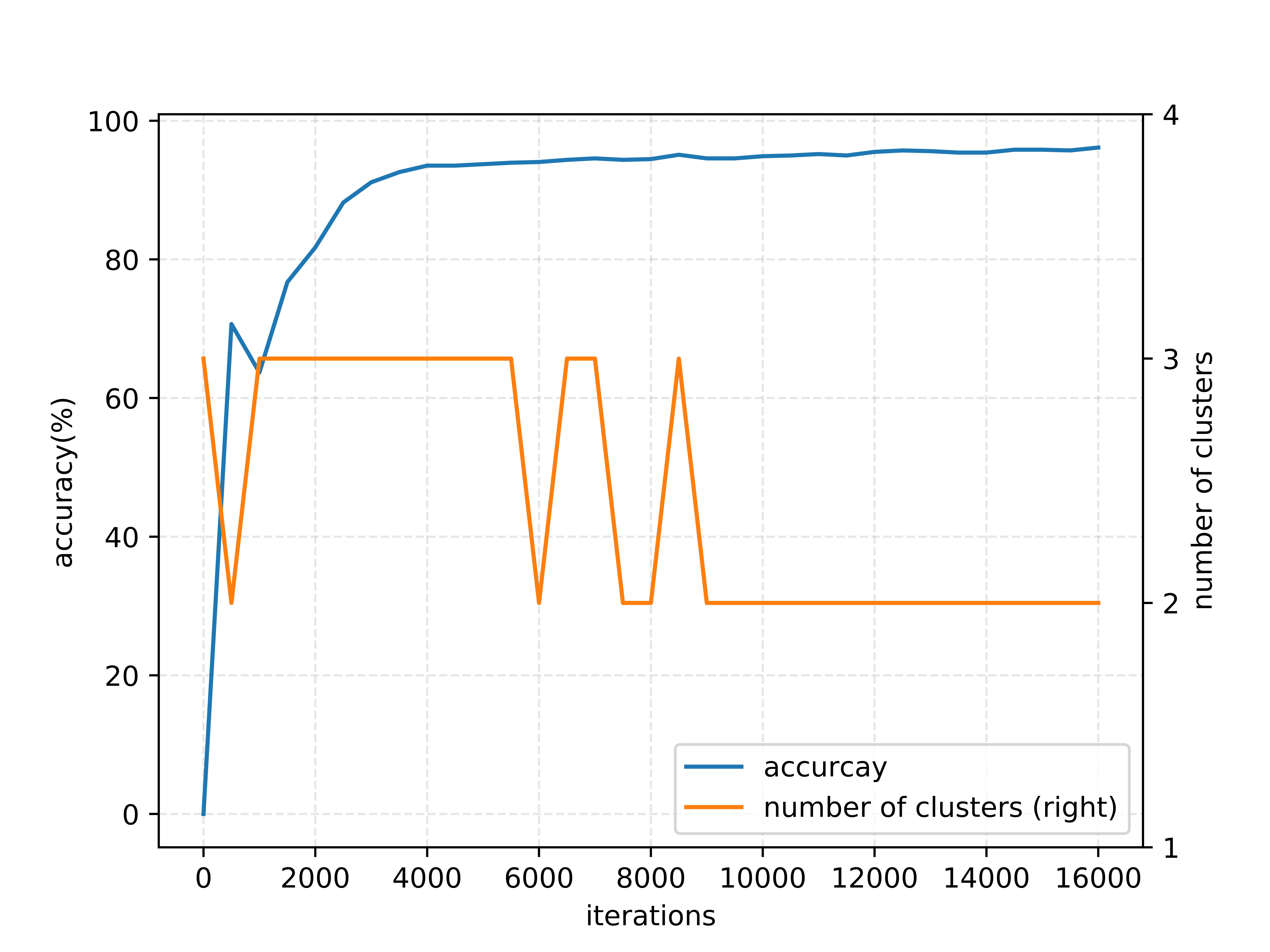}
\centering
\caption{The number of clusters w.r.t. to iterations}
\label{fig:1}
\end{figure}
\textbf{Feature Visualization}: 
Fig ~\ref{fig:t-sne}(a)-(d) are the feature visualization results by the t-SNE embeddings~\cite{Donahue} for ResNet, DaNN, PADA and SAPDA. The blue points, the green points and the red points represents source shared, source outlier and target samples respectively. Based on these four graphs, we can get some insightful observation: \textbf(1) ResNet-50 can only classifier a few target samples into correct categorises due to finetune, while DaNN almost cannot distinguish target samples into correct classes implies that mismatch between different label spaces deteriorate the accuracy. \textbf(2) PADA can cluster most target samples into correct classes but the bound between shared and outlier classes is not distinct, which can still cause misclassification. \textbf(3) Our SAPDA can not only distinguish target samples into correct classes, but also cluster source outlier classes together while the distance between other clusters is far. In this way, target samples can almost be misclassified.

\textbf{Number of Clusters}: Task W31 to A10 achieves the worst performance among the six tasks of Office-31 dataset, which means it is the most difficult task for our framework. Hence, in Fig.6 we utilize this task to shows the number of clusters w.r.t. to iterations. The left ordinate is the accuracy of the task w.r.t. to iterations, and the right ordinate is the number of cluster groups of the task w.r.t. to iterations.  

We have some interesting observations from this figure~\ref{fig:1} Even if we set the number of clusters ranging from 1 to 3, the actual number of clusters selected by the network only includes two or three under PDA setting. This phenomenon reflects that when it is a little hard to classify some mixed samples, the framework clusters our source data into three groups, while when the gap between shared and outlier classes is obvious, the framework can classify the shared and outlier classes easily, the number of cluster groups is two. \textbf(2) When the accuracy improves significantly, our network believes it has enough ability to handle the task, the number of cluster groups decreases from three to two. Once the accuracy decreases, the network can find out some of the classes having been misclassified, the number of cluster groups can increase from two to three. In this way, the misclassified classes can be adjusted until all the classes have been arranged into the correct groups. This phenomenon implies that our framework has the ability to self-adaptive correct weights for source classes.

\begin{figure}[tb!]
    \centering
    \includegraphics[scale=0.036]{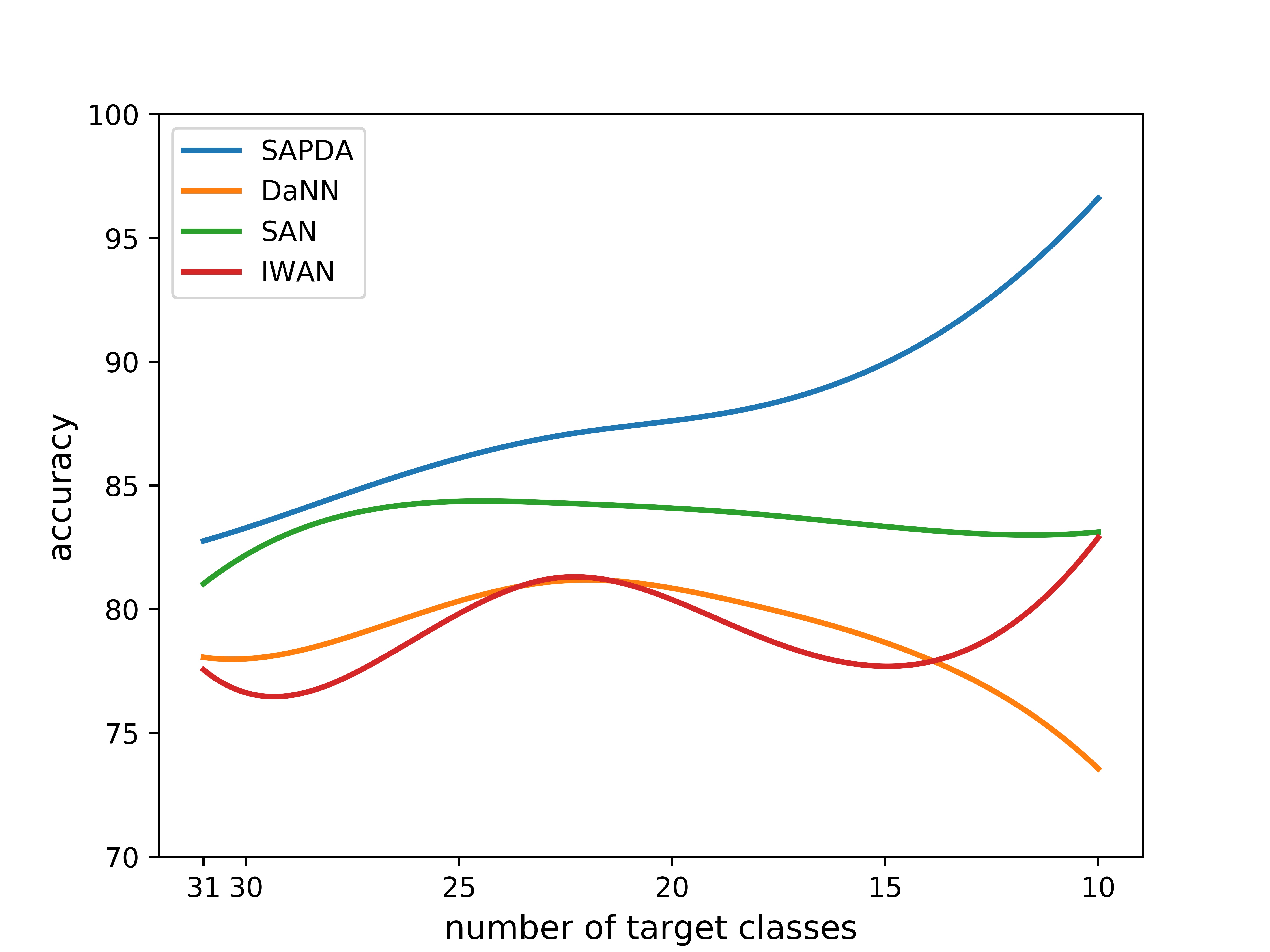}
    \centering
    \caption{Target test accuracy w.r.t. to target classes.}
    \label{fig:my_label}
\end{figure}

\textbf{Target Class}: We carry out some experiments with different target classes. Fig.7 shows DaNN performs worse as the number of target categories reduces, It indicates the influence of negative transfer caused by incongruity between different label spaces. Performance of SAN declines slowly and steadily, suggesting that the SAN has the potential to eliminate the effects of outlier classes. IWAN performs ordinary compared with SAN. Our SAPDA performs generally better than the other methods. Besides, when the number of target classes decreases, our model can achieve higher accuracy, showing our mechanism can not only select out outlier classes, but can also promote performance.
\begin{figure}
    \centering
    \includegraphics[scale=0.017]{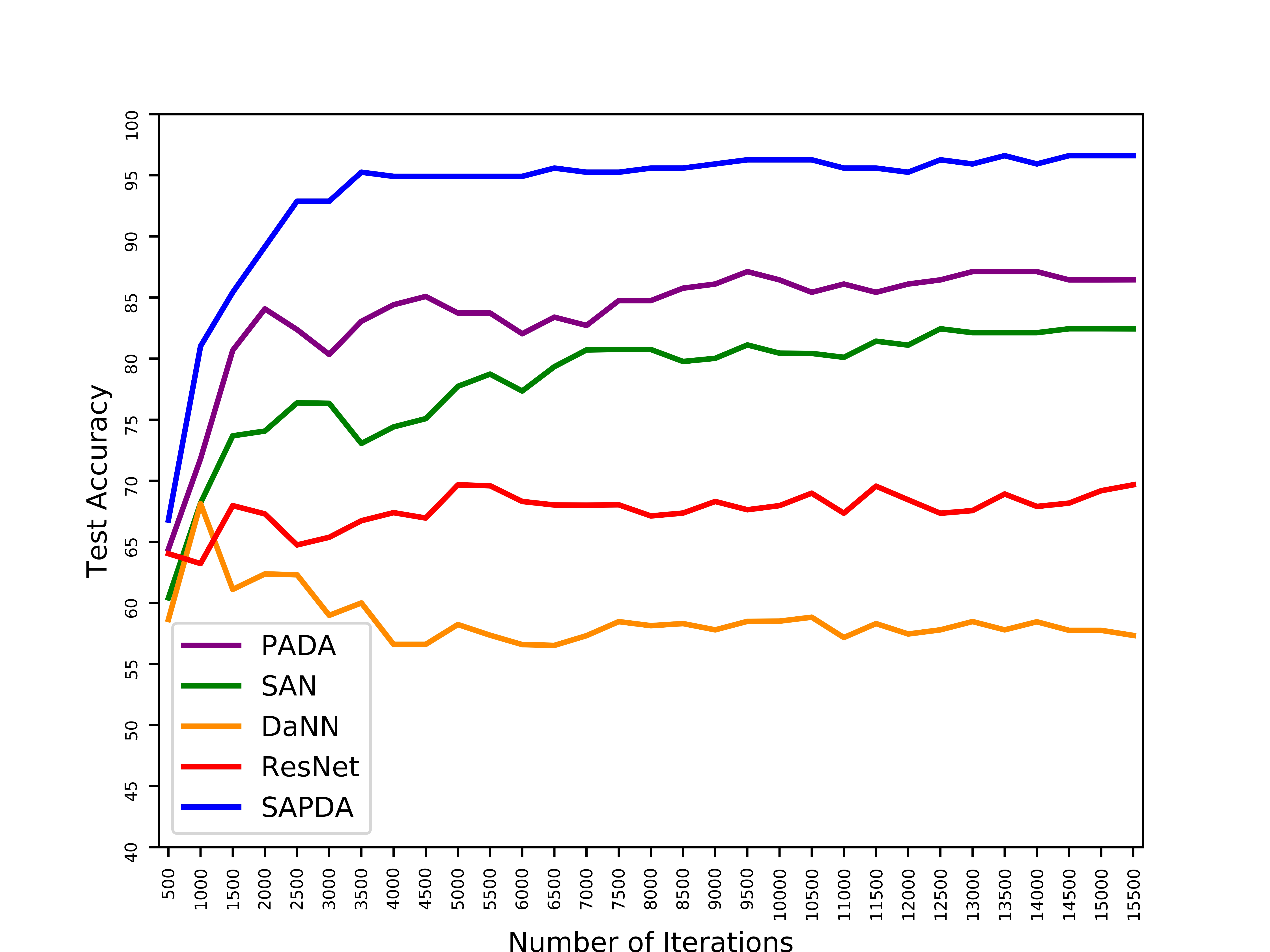}
    \centering
    \caption{Target test accuracy w.r.t. to iteration.}
    \label{fig:my_label1}
\end{figure}

\textbf{Sensitive Analysis:} In order to better analyze the sensitivity of our model to hyperparameter $\beta$, in Table VII, we observed the influence of different $\beta$ on experimental results on the Office-31 dataset. It is not difficult to find that the experimental results are best when $\beta=0.1$. Although other $\beta$ also affect the results, the overall results are relatively stable.

\textbf{Convergence Performance:} As shown in Fig.~\ref{fig:my_label1}, compared with it previous methods, our SAPDA does not only converge fast but also converges to highly accurate solutions, implying  the robustness and efficiency of our SAPDA.

\section{Conclusion} This paper presents an end to end Self-Adaptive Partial Domain Adaptation framework. It self-adaptively clusters source classes into different groups, and samples in the same group having the same weights. In this way, weighted adversarial network progressively quantifies the transferability of source examples, and simultaneously learns domain-invariant features across source and target domains. Experiments show effectiveness of our model and superiority over several benchmarks.


%





\ifCLASSOPTIONcaptionsoff
  \newpage
\fi



%
\bibliographystyle{plain}
{
\bibliography{egbib}}
\end{document}